# Milestone Determination for Autonomous Railway Operation.

Josh H* · John McD · Simon B · Poppy F · Mia D

**Abstract** In the field of railway automation, one of the key challenges has been the development of effective computer vision systems due to the limited availability of high-quality, sequential data. Traditional datasets are restricted in scope, lacking the spatio-temporal context necessary for real-time decision-making, while alternative solutions introduce issues related to realism and applicability. By focusing on route-specific, contextually relevant cues, we can generate rich, sequential datasets that align more closely with real-world operational logic. The concept of milestone determination allows for the development of targeted, rule-based models that simplify the learning process by eliminating the need for generalized recognition of dynamic components, focusing instead on the critical decision points along a route. We argue that this approach provides a practical framework for training vision agents in controlled, predictable environments, facilitating safer and more efficient machine learning systems for railway automation.

## 1. Background & Positioning

Driverless metros have operated since the 1920s [10], however, innovation within the railway sector has slowed down in recent years. Railway automation has advanced unevenly across domains. Driverless metro systems (GoA-4) have expanded where infrastructure is closed and controlled but translating that success to open mainline environments has proved difficult, largely because perception and assurance problems differ fundamentally from fenced metros and other domains such as road autonomy. Standards and surveys consistently note both the appetite for automation and the gap in practicable guidance for Machine Learning (ML) based perception and control in rail. Peleska et al. highlight that prevailing CENELEC standards provide no concrete pathway for ML components in autonomous train control, underlining why research often remains speculative when applied to safety-critical rail operation [18, 21].

A broad literature survey echoes that AI in rail has focused more on maintenance, scheduling, and diagnostics than on ego-vision for autonomous operation; truly end-to-end Automatic Train Operation perception remains niche [3]. Our previous research interviewing railway professionals suggests that Computer Vision (CV) techniques are not directly transferable between cars and trains [8]. However, that has not stopped attempts at developing railway systems based on State of the Art (SotA) automotive technologies [1 , 22]. Fundamentally, automation within the railway space has been agreed upon using the Grades of Automation (GoA) classification [20], however, there have been disagreements with the classification, e.g., Professor Tang Tao from Beijing Jiaotong University has proposed a separate classification of automation which more reflects the level of responsibility placed onto the autonomous vehicle itself [24, 25, 29].

This paper presents an approach to addressing some of the problems of high levels of autonomy within rail, regardless of how it is classified. Specifically, it operationalises part of the SACRED methodology [7], which gives a structured approach to the assurance of autonomous railway systems. Specifically, we focus on representing a route through an Operational Domain Model (ODM), using milestones as context-dependent anchors.

One study considering context-sensitive information within rail is the study "Hidden Markov Random Fields an application to railway" by Come et al [39] modelling both sensor observations and spatial label patterns of track features. This approach demonstrated that incorporating context improved defect detection accuracy over classifiers using sensor data alone. However, Come's work was regarding modelling in diagnosis tasks, rather than for real-time operational decision-making. Moreover, their performance relied on dense labelled datasets which are not widely available. The remainder of this paper demonstrates how we modelled this structure in simulation, defined contexts with Human-in-the-Loop input, and evaluated predictive performance against traditional probabilistic classification for operational decision-making.

### 1.1 Dataset Limitations

When considering milestones and the approach we will use to demonstrate their potential, we compare contextual information to standard information using deterministic and probabilistic methods, this is because traditional ML approaches are limited via data availability. A study by Moloney highlights that SotA railway simulations offer scarce data with high acquisition costs [16]. Existing CV datasets for railways exhibit notable gaps in temporal and contextual information, despite recent progress in visual annotation. Both RailSem19 [28] and OSDaR23 [23] offer high-levels of ego-mounted data, however they do not offer the rest of the ecosystem-level information, such as signal states or timetabling information, required for a train to decide how to operate. As Medeossi and Fabris note no comprehensive "all-in-one" simulator for rail exists [14]. To gauge practical detectability at mainline stopping distances, we performed a minimal check using OSDaR23. As suggested within Tagiew's paper [23], we used the available devkit raillabel (https://github.com/dbinfrago/raillabel) before converting the labelled set to COCO format for the purposes of YOLOv11 deployment [9]. From here we captured some UK mainline data using 2K cameras at 400m from green lineside signals. In our previous work, we argued that safety analysis for autonomous rail systems must identify "measurable attributes" that map directly onto operational constraints [7]. One such attribute is the requirement that signals be perceivable within a 4–8 second window before passing, which at UK mainline speeds (125 mph / 55 m·s) corresponds to 220–440 m. This standard is embedded in UK driver handbooks as a buffer against Signals Passed At Danger (SPAD). We therefore adopted 400 m as a representative test distance and 2K imagery as a proxy for contemporary high-end rail-mounted cameras. At this distance, OSDaR23 failed to produce a valid detection. This is unsurprising as OSDaR23 uses image resolutions ranging from 240p to 1200 and has emphasis on obstacle/track/distance tasks rather than long-range signal recognition; small, low-contrast targets at 200–400 m occupy only a handful of pixels and are easily dominated by background clutter.

* Corresponding author : josh.hunter@york.ac.uk

Limitations in data availability within rail has been cited as a key reason behind the usage of less-than-optimal datasets, data collected within videogames such as Grand Theft Auto V [12], data collected by hand [19] or datasets which resulted in underwhelming results [4, 5, 15]. Within previous work, we explored the information required by cab drivers for safe railway operation and found that railway safety is largely defined by the interaction between drivers and signal operators [8], suggesting visual information alone is not enough to make a full decision. ETCS systems communicate contextual information through radio communication [17], Tetsuo Uzuka's 'system of systems' model discusses the full web of contextual information required for proper communication [27]. CV datasets are valuable, but all pursue a universal framing of perception; all with the underlying assumption that a single, universal dataset can support all perception needs. We reject this viewpoint; railway safety depends on context. What matters to a driver approaching a station differs from what matters when cruising at line speed or entering a tunnel. We call this contextual information 'milestones.'

### 1.2 Humans in the Loop

While milestones provide the structural information for determining when transitions between operational states occur, they do not in themselves prescribe the quantitative parameters that govern safe behaviour within those states. In other words, the appearance of a platform edge signals that Cruise should transition into Station Approach, but it does not specify the precise deceleration profile, the margin of braking force to be applied, or the safety limits on speed reduction. This distinction between when to transition and how to parameterize the subsequent state is critical. Within the SACRED methodology, safety attributes only acquire meaning when contextualized in time and space; hazards shift with each mode, and the appropriate numerical thresholds must be calibrated accordingly.

To account for this requirement for contextualization, we design our metrics with a Human in the Loop (HitL). Lloyd-Roberts et al. demonstrate how interactive, HitL invariant finding can be used to improve railway safety, showing the value of co-design between engineers and automated verification tools [11]. Similarly, Thron et al. highlight that human oversight remains indispensable for resilience in railway cyber-security, as automated systems alone cannot adequately address emergent risks in complex socio-technical environments [26]. For this specific experiment, the HitL provided context on which variables were most important within a given state, these variables were then fine-tuned to provide optimal results. In summary, Milestones define when transitions occur. HitL define how behaviour within those states should be parameterised.

## 2. Representing the Operational Domain Model through Milestones

When quantifying the operation of an autonomous system (AS), common practice is to first design the parameters in which the AS will operate. Within top-down systems design, this takes the form of an Operational Design Domain (ODD) [1,32]; we have previously argued that within rail we instead should quantify systems using the Operational Domain Model (ODM) [8]. A key difference between the ODM and the ODD is the framing of hazards, where an ODD quantifies exactly what hazards a given system can interact with, the ODM quantifies exactly which hazards any system can expect to encounter within a given environment [30,31].

* Corresponding author : josh.hunter@york.ac.uk

A core premise of this work is that, unlike the open road domain, the railway is a well-structured and repeatable environment. Except for adjustments such as weather, lighting, or incidental obstacles, the spatial configuration of infrastructure along a given line remains fixed. This allows safety analysis to be scoped on a by-route basis rather than seeking a universal, one-size-fits-all solution. Hong et al. argue that looking at rail with the philosophy of systems thinking-based theory of accident causation means that safety analysis requires situating hazards and defences in their route-specific system context, rather than attempting to generalise across the entire network [6]. Where Hong presents this as a weakness, suggesting that systems-based thinking is therefore not scalable, we embrace the philosophy, arguing that any autonomy within rail must be specific to the given route, much like how a given route must be learned by a driver prior to them driving it operationally. Specific portions of track require different approaches to driving, when approaching a station, a train's objective is to stop (assuming this is part of the schedule), when approaching a green signal, a trains objective is to maintain speed.

The value of state-based modelling in railway systems is well established. Benerecetti et al [40] argue that railway control is inherently sequential and rule-driven, which is why international standards EN 50129 [2] recommends state machines as a verification tool, Benerecetti propose a Dynamic State Machine to properly represent communication, concurrency, and timing; we are arguing that these transition semantics can be achieved through milestones. This interpretation aligns with safety assurance principles. Our own state diagram (Figure 1) represents a simplified model of this principle. We argue that a state machine is a simplified expression of the environment, or a simplified ODM.

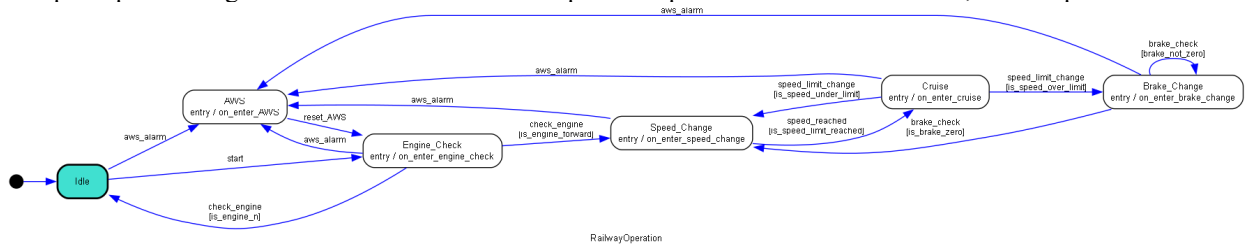

**Fig. 1.** A simplified ODM represented as a state machine. [41]

Safety attributes are only meaningful when contextualised in time and space, with hazards and safe operating limits shifting as the system transitions between modes or states. McLeod et al. highlight that train drivers' performance depends critically on their situational model of "now," which is continuously updated through route knowledge and environmental cues [13]. In essence, drivers already navigate railways as a sequence of states, transitioned by milestones. If a state machine is representative of the environment, then milestones are the transition between states.

## 2.1 The Observed weight of an Output

We have argued that in order to properly represent railway operation, contextual information must be included. To operationalise this concept, we required a controllable environment in which states could be explicitly annotated and manipulated. For this purpose, we adopted OpenBVE [3], a ferroequinologist speciality simulator. OpenBVE allows for the full simulation of a railway route from the perspective of a cab driver, evaluating performance with an internal score. For experimentation, we wanted to evaluate two reinforcement learning strategies, one using Gaussian Naïve Bayes which considers all variables to be independent of one another, one using a modified Bayesian framework which has individual weighting based on contextual information, we achieve this by using the formula:

$$\hat{I} = \arg\max_{I} \left( P(I) \prod_{o \in O} P(o \mid I)^{wO} \right)$$

Here $P(I)$ represents the prior probability of the input, $P(o \mid I)$ represents the conditional probability of observing $o$ given $I$. $wO$ is the weight of the output $O$ as determined by the current state. The specific values of $wO$ were determined by a HitL, a railway professional who was familiar with the given route [8]. Context was explained in a conversational setting and $wO$ were fine-tuned for optimal results, allowing us to algorithmically predict the next input by Observing the weight of the previous Output (OwO). To evaluate this framework, we conducted 25 simulated runs of a trial route (from Swalwell Junction to Newcastle) in OpenBVE, collecting over 2M datapoints including speed, braking, acceleration, and signal data. A HitL railway professional assigned state-specific weights, which we then tested by comparing Gaussian Naïve Bayes to our OwO model.

* Corresponding author : josh.hunter@york.ac.uk

Within OpenBVE, the possible inputs are Power (*P*) and Brake (*B*), both of which are represented as *PI*. The contextual information given by the simulator are Time (*T*), Speed (*S*), Speed Limit (*SL*), Rate of Acceleration (*RoA*), Signal Limit (*SLS*) and Engine State (*ES*). Within different 'states' the system has a different set of given goals, for instance, in the 'Cruise' state, the train's primary objective is to maintain $S < SL$. In this context, the *SL* is the most critical observed output, while the *RoA* becomes less relevant since maintaining speed takes precedence over rapid changes. Conversely, in the 'Brake_Change' state, where a significant reduction in speed is required, *RoA* gains critical importance as the system needs to decelerate effectively without risking a complete stop or violating safety thresholds. The *OwO* framework dynamically assigns weights to outputs based on their relevance within each state, ensuring that predictions about the next inputs are context-sensitive and optimized for the current operational mode. Within Gaussian Bayes, each input to the system is weighted independently from one-another, so, they are each weighted at 100%, using OwO however, each input has a different weight depending on the context, or state of operation. The specific representation of each states weighting is shown in Table 1.

**Table 1.** Observation Weights Breakdown by State

| State | T | S | SL | SLS | RoA | ES | PI | Description |
|---|---|---|---|---|---|---|---|---|
| Cruise | 135 | 150 | 150 | 0 | 10 | 0 | 0* | Maintains speed close to the limit between stations. Adjustments are minimal and handled in Brake_Change or Speed_Change. |
| AWS | 0 | 0 | 0 | 200 | 0 | 0 | 200* | Triggered when passing a signal; train exits its current state to acknowledge AWS. Only valid input: setting both *P* and *B* to 0. |
| Engine_Check | 0 | 0 | 0 | 0 | 0 | 200 | 0* | A buffer state at journey start or after AWS exit, ensuring $\hat{I} = 0$ before speed modifying states. |
| Brake_Change | 10 | 150 | 150 | 50 | 150 | 0 | 50* | Reduces speed to stay within limits. Considers *S*, *SL*, and *RoA* while avoiding counteracting inputs |
| Speed_Change | 10 | 150 | 150 | 50 | 150 | 0 | 50* | Increases speed after a drop. Similar to Brake_Change but ensures consistency and prevents over-correction. |

Table 2 shows a comparison between Gaussian Naive Bayes (NB), and two versions of the *OwO* model described within the arg-max formula. One model with previous inputs excluded and one with them included. All models aim to predict the next input sequence $\hat{I}$ based on observed outputs O. We compare *OwO* to Gaussian NB due to the similarity between the two, to determine if providing additional context results in an improvement of accuracy.

**Table 2.** Comparative Accuracy Results Between Naive Bayes and Observed Weight of Outputs Models

| State | NB Acc. | OwO W/O PI | OwO with PI | Description |
|---|---|---|---|---|
| Cruise | 65-75 | ~80% | ~90% | Failures in models typically occur due to over-speeding. |
| AWS | ~99% | ~99% | ~99% | AWS state has a single valid input: setting both *P* and *B* to 0. The small error margin (~1%) is due to minor delays leading to slight inaccuracies in frame-perfect inputs. |
| Engine_Check | ~65% | ~60% | ~85% | NB outperforms OwO in transitioning into *EC*. When previous inputs are given alongside the context of transitioning state, OwO's accuracy improves substantially. |
| Break_Change | ~85% | 95-99% | ~82-99% | Ow*OPI* struggles due to undercorrection when *P* and *B* are low, however Ow*OPI* is quicker than NB when *P/B* are high |

* Corresponding author : josh.hunter@york.ac.uk

| | | | | |
|---|---|---|---|---|
| Speed_Change | ~85% | 95-99% | ~82-99% | Similar to Brake Change, OwO outperforms NB by effectively handling input adjustments, preventing simultaneous or conflicting changes |

Overall, the OwO model improved accuracy by ~5% across most states, particularly in Cruise, Brake_Change, and Speed_Change, where contextual awareness is critical. When the *PI* is introduced, there is further improvement. However, an analysis of SotA has proven difficult for the capturing and remembering of PI within a non-simulated environment, while OpenBVE has the memory capacity for capturing previous states, our experiments outside of OpenBVE have struggled to replicate this, which we discuss further within the section on future work.

# 3. Conclusion

This study operationalises SACRED by showing how ODMs can be represented through state machines, milestones used as transitions, and operational contexts quantified through OwO weighting. Our findings suggest that the incorporation of contextual information significantly enhances performance in states where decision-making relies on a nuanced understanding of the operational environment. This aligns with real-world railway training, where operators are taught to adapt to route-specific and situational factors. However, the route-specific nature of $OwO$ highlights the need for robust ODM definitions to generalize the approach across various routes and scenarios.

While further research is required to refine the definition and implementation of contextual weights, this study demonstrates the potential of the $OwO$ model as a foundation for supervised learning in railway simulation. The results are promising and underscore the value of integrating state-dependent observations into predictive frameworks for railway automation.

## 3.1 Future Work

Within this work we have introduced milestones as a method to structure and assist in the classification of railway operation. On their own, milestones are abstract, but the $OwO$ model demonstrates one way they can be translated into machine-readable metrics that improve state-specific performance. A key open question is how a system can robustly determine that it has crossed a milestone outside of simulation. In our experiments, simulation environments made this straightforward. We could track the absolute cab position and maintain a perfect memory of state, enabling a supervised learning setup that reacted reliably to contextual information. Translating this approach to real-world systems, however, remains challenging, as SotA railway perception platforms rarely include the kind of robust memory functionality required to mimic the *PI* variable used in our experiments. Part of the SACRED methodology is analysing how the SotA lines up against the requirements determined by the ODM. To this end, we aim to find out how the SotA can determine and handle Milestones.

### 3.1.1 Visual Milestone Detection

One avenue for the implementation of milestones within machine learning systems is through context-sensitive CV. Work on few-shot classification with temporal alignment [36] has suggested that supportive datasets can help model context within a scenario that is commonly encountered although early results have been mixed due to data availability constraints. Extending this line of work with richer, route-specific datasets would help to ground milestones in real operational contexts. However, the problem simply returns to dataset availability, current SotA datasets [23,28] lack sufficient contextual information to work as the basis of milestone determination, one potential avenue for Milestones within CV is the use of trackside cameras, however, the usage of this technology is sparsely developed [37].

### 3.1.2 Reinforcement Learning using Milestones

A further strand of future work lies in embedding milestones within reinforcement learning (RL) frameworks. Conventional RL approaches typically rely on exponential discounting of future rewards, which biases decision-making toward short-term gains and undervalues events that occur further in the future. For autonomous railway systems, this is problematic. Many of the most safety-critical events, such as signals, station stops, or level crossings, are defined as milestones that may only materialise after several minutes of operation. An agent optimising under exponential discounting may therefore neglect the influence of distant but vital milestones in favour of immediate corrections.

* Corresponding author : josh.hunter@york.ac.uk

Schultheis et al. have shown that non-exponential discounting functions better to capture human decision-making, where long-term goals and obligations retain salience even when temporally distant. By adapting this principle, we propose to integrate milestones as anchors in RL reward structures, and to employ non-exponential discounting so that these anchors continue to shape short-term policy. Initial experimentation of RL-milestones has shown encouraging potential, with agents retaining sensitivity to distant operational events while still responding effectively to short-term fluctuations. However, this line of work remains in an exploratory stage, additional refinement of the reward structure and more extensive validation across varied routes are required before firm conclusions can be drawn.

### 3.2 Closing Remarks

Current approaches to railway automation rarely consider the wider ecosystem in its entirety when designing systems, it is also clear that when designing a railway system, oftentimes the designer of the cab has little control over the ecosystem itself, with signalling systems and transmission of information often being pre-determined. Within our wider scope of work, we argue that rail should be considered an ego-vehicle within an ecosystem; the cab is an autonomous entity which makes decisions based on information given by external actors. Milestones, therefore, are a codification of this information in a way that "makes sense" to a machine learning system. Previous studies have discussed how different domains require unique approaches to quantify the environment [33,34.35], this study has introduced milestones as a machine learning focused approach to designing railway ODM's within the framework of the SACRED methodology, in order to assist in the development of railway-specific technologies.

* Corresponding author : josh.hunter@york.ac.uk

* Corresponding author : josh.hunter@york.ac.uk

## Author List

Josh Hunter : josh.hunter@york.ac.uk - 0000-0002-6828-974X
   University of York, Centre for Assuring Autonomy, York, United Kingdom
John McDermid: john.mcdermid@york.ac.uk
   University of York, Centre for Assuring Autonomy, York, United Kingdom
Simon Burton; Simon.Burton@york.ac.uk
   University of York, Centre for Assuring Autonomy, York, United Kingdom
Poppy Fynes: pf693@york.ac.uk
   University of York, Centre for Doctoral Training in Safe Artificial Intelligence Systems (SAINTS) (EP/Y030540/1)', York, United Kingdom
Mia Dempster: - 0009-0009-3035-7303
   Leeds Law School, Leeds Beckett University, Leeds, United Kingdom

## Declaration

This work is undertaken with support from Siemens mobility, and in conjunction with the Fraunhofer Institute
for Cognitive Systems (IKS) in Munich. This work was undertaken with the cooperation of the Associated Society of Locomotive Engineers and Firemen (ASLEF). Some of the co-authors of this work are supported by UKRI AI Centre for Doctoral Training in Safe Artificial Intelligence Systems (SAINTS) (EP/Y030540/1)'

* Corresponding author : josh.hunter@york.ac.uk